\documentclass[acmtog]{acmart}
\AtBeginDocument{%
  }

\usepackage{xspace}
\usepackage{enumitem}
\usepackage{hyperref} 
\usepackage{cleveref} 
\usepackage{xcolor,colortbl}
\usepackage{comment}
\usepackage{pifont}
\usepackage{tikz}
\usepackage{xspace}
\usepackage{bbding}
\usepackage{caption} 
\usepackage{float}
\usepackage{subcaption}
\usepackage{url}
\usepackage[linesnumbered,ruled,vlined]{algorithm2e}
\usepackage{color,multirow,microtype} 
\usepackage{booktabs}
\usepackage{amsthm} 
\usepackage{pifont}

\crefname{theorem}{Theorem}{Theorem}
\crefname{lemma}{Lemma}{Lemma}
\crefname{remark}{Remark}{Remark}
\crefname{figure}{Fig.}{Fig.}
\crefname{section}{Sec.}{Sec.}
\crefname{equation}{Eq.}{Eq.}
\crefname{table}{Tab.}{Tab.}
\crefname{algocf}{Alg.}{Alg.}

\copyrightyear{2026}
\acmYear{2026}
\setcopyright{cc}
\setcctype{by}
\acmConference[SIGGRAPH Conference Papers '26]{Special Interest Group on Computer Graphics and Interactive Techniques Conference Conference Papers}{July 19--23, 2026}{Los Angeles, CA, USA}
\acmBooktitle{Special Interest Group on Computer Graphics and Interactive Techniques Conference Conference Papers (SIGGRAPH Conference Papers '26), July 19--23, 2026, Los Angeles, CA, USA}
\acmDOI{10.1145/3799902.3811062}
\acmISBN{979-8-4007-2554-8/2026/07}

\definecolor{lightred}{RGB}{255,150,150}
\definecolor{lightblue}{RGB}{150,150,255}
\definecolor{darkred}{RGB}{255,150,150}
\newcommand{\methodname}{\textsc{Motion4Motion}\xspace}
\newcommand{\tpe}{\textsc{TransPE}\xspace}

\begin{document}

\title{\methodname: Motion Transfer Across Subjects at Inference}

\author{Ling-Hao Chen}
\email{evan@lhchen.top}
\affiliation{%
  \institution{Tsinghua University}
  \country{China}
}
\affiliation{%
  \institution{Stepfun}
  \country{China}
}

\author{Zixin Yin}
\affiliation{%
  \institution{The Hong Kong University of Science and Technology}
  \country{China}}
\affiliation{%
  \institution{Stepfun}
  \country{China}
}

\author{Duomin Wang}
\affiliation{%
  \institution{Stepfun}
  \country{China}
}

\author{Xianfang Zeng}
\affiliation{%
 \institution{Stepfun}
 \country{China}}

\author{Gang Yu}\authornote{Corresponding author.}
\affiliation{%
  \institution{Stepfun}
  \country{China}}

\renewcommand{\shortauthors}{Chen et al.}

\begin{abstract}
  This work explores the motion transfer from one video to another, which is crucial in animation for diverse characters. Previously, video motion transfer has been largely explored between human and human-like characters, enabling a lot of applications in digital creation. However, these approaches encounter a main limitation. Specifically, related technical pipelines heavily rely on a predefined human skeleton structure and accordingly require skeleton-conditional model training. On the one hand, these methods are difficult to generalize to diverse characters, such as animals from different species, while preserving their unique motion styles. On the other hand, labeled data in diverse skeletons is limited, which additionally restricts the large-scale training for the task. In this paper, we jump out of the skeleton-based motion transfer framework and propose a training-free motion transfer framework, named \methodname. \methodname models the motion flow of the character in a video instead of skeletons, which makes motion transfer across species easier. Extensive experimental results and novel applications show our methods outperform baselines impressively. Project page is available at \url{https://lhchen.top/Motion4Motion/}. 
\end{abstract}


\begin{teaserfigure}
  \includegraphics[width=\textwidth]{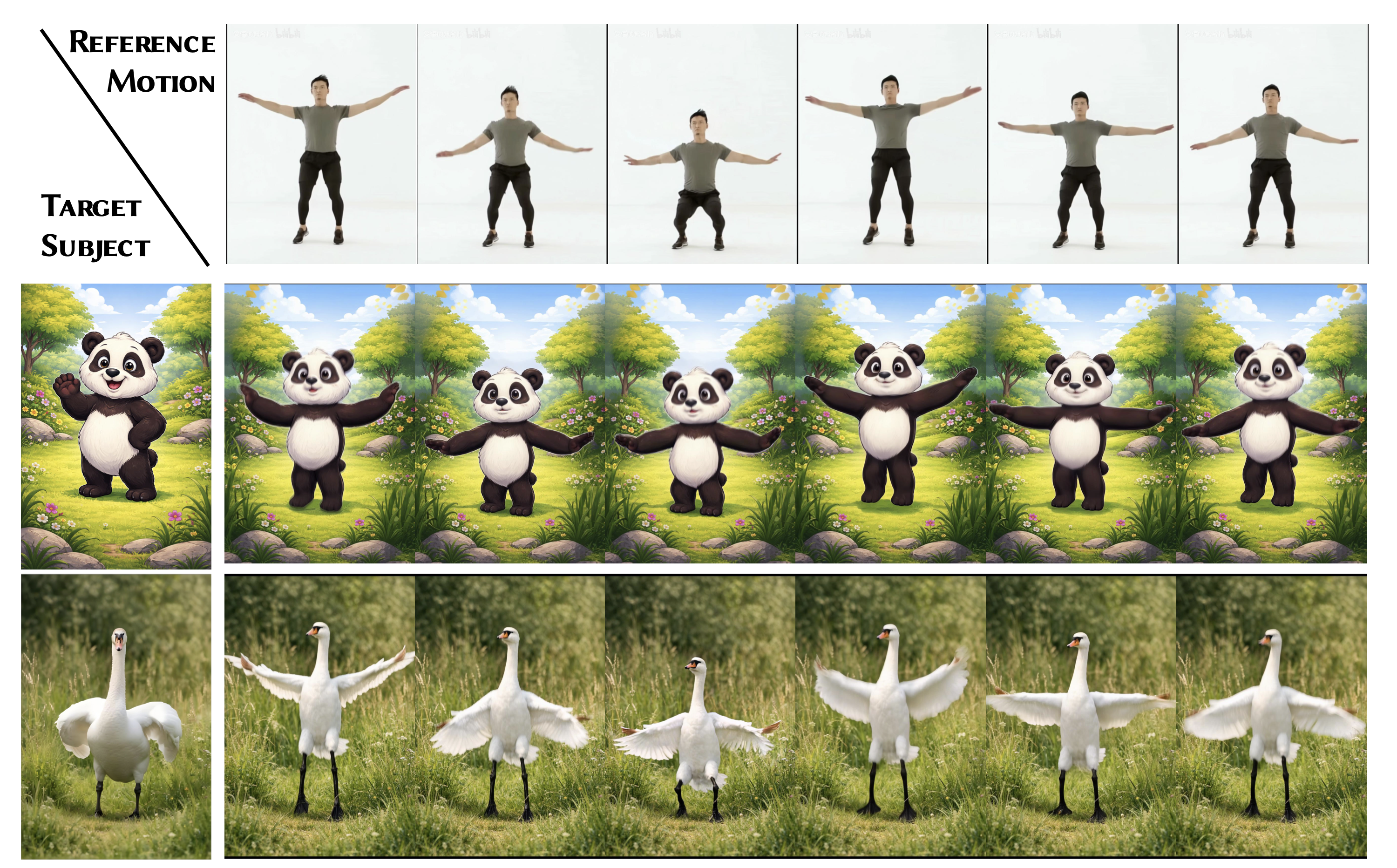}
  \caption{\textbf{Introducing \methodname, a framework transferring motion from one subject to another.} \methodname achieves cross-species (\textit{e.g.}, human $\to$ panda, or human $\to$ goose) motion transfer without a uniform skeleton at inference.}
  \label{fig:teaser}
\end{teaserfigure}

\maketitle

\section{Introduction}
\label{sec:intro}

Motion transfer~\cite{hu2024animate,guoanimatediff,zhang2025flexiact} has a wide application in digital creation and animation workflows, such as character animation~\citep{chen2025motion2motion}, virtual reality, and movie post-production. In recent years, the mainstream research has primarily focused on human-centric scenarios, where the core objective is to migrate movements from a source person to a target person. To achieve this, existing methodologies heavily rely on skeletal representations to bridge the geometric gap between different subjects. By extracting structural poses, these skeleton-based pipelines have achieved remarkable success in transfer quality. 

Despite these progresses, existing frameworks face significant hurdles in real-world applications, particularly when extending beyond human-centric domains. Most current paradigms are intrinsically ``hard-coded'' for specific structural priors, which severely restricts their flexibility across different species. When tasked with transferring motion between characters with vastly different morphologies, such as among various animal species, the lack of a shared skeletal template makes spatial alignment ill-defined. This rigidity prevents current methodologies from generalizing to ``in-the-wild'' scenarios where characters may possess arbitrary shapes and motion styles that deviate significantly from standard human proportions. Notably, even the most relevant attempt, FlexiAct~\cite{zhang2025flexiact}, remains unsatisfactory, as it relies on per-case optimization that leads to overfitting and consequent information leakage.

Two fundamental challenges impede the realization of robust, cross-species motion transfer in a more general space. The first challenge is the critical scarcity of high-quality, paired motion data across diverse topologies. Unlike human-centric research, which benefits from massive video datasets and mature pose estimation tools, obtaining synchronized motion sequences for diverse characters is both labor-intensive and often impractical. This data bottleneck forces data-driven models to rely on narrow distributions, leading to severe artifacts when encountering unseen species. The second challenge involves the ambiguity of defining semantic correspondences between source and target subjects without skeletons, such as ambiguity between the legs of a chair and a quadrupedal animal. Establishing consistent mappings becomes exceptionally difficult when the target character possesses completely different physical semantics, making it hard to maintain motion fidelity while ensuring the visual plausibility of the transferred results. 

To overcome the inherent constraints of predefined skeletal templates in driving novel character animations, we depart from the conventional motion transfer pipeline and propose \methodname, a novel framework designed for video-based motion synthesis. Unlike previous kinematics-based methods that rely on skeletal priors, \methodname operates on dense pixel-level motion flows, treating them as the fundamental primitives for motion representation. By capturing the temporal dynamics of source pixels and mapping them onto the target subject via our proposed \tpe module, our method achieves high-fidelity transfer without structural limitations. Notably, \methodname is a training-free approach implemented entirely during inference, which enables the good interpretability. 

Before delving into details, our core contributions are as,
\begin{itemize}
    \item We present \methodname, the first training-free framework capable of transferring motion across general subjects without relying on predefined skeletal priors.
    \item We propose a simple yet effective module, namely \tpe, for injecting motion flows to the target subject. 
    \item Extensive evaluations demonstrate that \methodname not only delivers high-fidelity motion transfer results but also exhibits potential applications in novel concept composition and cross-morphology motion transfer (\cref{sec:app}). 
\end{itemize}

\section{Related Work}
\subsection{Video Generation}
While early diffusion models predicated on U-Net backbones~\citep{ho2020denoising, rombach2022high, guoanimatediff} significantly outperformed GAN-based systems~\citep{reed2016generative, yu2023talking, wang2023progressive} in image fidelity, their scalability constraints have precipitated a paradigm shift toward Diffusion Transformers (DiTs)~\citep{peebles2023scalable, esser2024scaling,wang2025universe}. DiTs have since emerged as the foundational backbone for state-of-the-art video diffusion models, exemplified by CogVideo-X~\citep{yangcogvideox} and WAN~\citep{wan2025}. In this work, we propose a novel attention control method designed to integrate with the WAN.

\subsection{Attention Control}
Parallel to these architectural advancements, the field of controllable generation has expanded significantly. Originating with Prompt-to-Prompt~\citep{hertzprompt}, attention control methods have been extensively deployed to modulate pre-trained U-Nets for image and video editing tasks~\citep{tumanyan2023plug, cao2023masactrl, liu2024video, yin2025consistedit, yin2025lazydrag,yin2025training}. In particular, key-value (KV) injection and concatenation in attention layers have been explored for various purposes, including style-consistent generation~\citep{hertz2024stylealigned}, zero-shot style transfer~\citep{deng2024zstar}, subject-driven consistent generation~\citep{tewel2024trainingfree}, and appearance transfer~\citep{alaluf2024crossimage}. Although recent endeavors have begun extending these controls to DiTs~\citep{yin2025consistedit, wang2024taming, cai2024ditctrl, yin2025lazydrag}, they predominantly target MM-DiT architectures~\citep{esser2024scaling}, which rely on a unified self-attention mechanism for fusing visual and textual modalities. Consequently, the efficacy of attention control within DiT architectures employing \textit{decoupled} self-attention and cross-attention layers, such as WAN, remains underexplored. Furthermore, while existing literature addresses general editing and long-video synthesis, the potential of DiT-based attention control specifically for motion transfer has yet to be investigated.

\subsection{Motion Transfer}
\paragraph{3D-based motion transfer.}
The problem of motion transfer between different characters was first established and extensively explored in the 3D animation community as motion retargeting~\citep{gleicher1998retargetting}. Traditional methods relied on kinematic optimization to satisfy spatial constraints~\citep{feng2012automating, lee1999hierarchical}, while modern neural-based approaches~\citep{aberman2020skeleton, lim2019pmnet} leverage deep architectures to decouple pose and structure. Although some recent iterations~\citep{li2022iterative, chen2025motion2motion} attempt to improve generalization across disparate morphologies, they remain bound to 3D skeletal topologies~\citep{lu2023humantomato,chen2023humanmac,chen2024pay,motionlcm} and often require manual joint correspondences. Transitioning these concepts to the video domain presents unique challenges, as explicit structural information is often absent or noisy in raw pixels.

\paragraph{Video-based motion transfer.} 
Specialized human animation frameworks~\citep{hu2024animate, zhangmimicmotion, cheng2025wan} built upon pre-trained video diffusion models rely heavily on large-scale pre-training and explicit skeletal guidance, which limits their applicability to fixed topologies and demands massive computational resources. Conversely, general motion editing approaches like FlexiAct~\citep{zhang2025flexiact}, and others~\citep{zhao2024motiondirector, lingmotionclone, burgert2025motionv2veditingmotionvideo, gokmen2025ropecraft} avoid skeletal constraints but typically necessitate time-consuming per-video fine-tuning. 
Motion Prompting~\citep{geng2025motionprompting} proposed ControlNet-based training for general trajectory-guided motion control. Go-with-the-Flow~\citep{burgert2025gowiththeflow} introduces warped noise for real-time motion-controllable generation. Diffusion-As-Shader~\citep{gu2025diffusionshader} and WAN-Move~\citep{chu2025wanmove} train dedicated trajectory-conditioned modules for video control. ATI~\citep{wang2025ati} further unifies trajectory instructions for controllable generation, while MotionStream~\citep{shin2026motionstream} achieves real-time interactive motion control. Despite the impressive performance, these training-based methods are inherently limited by the coverage of their training data and may not generalize to unseen motion patterns. In contrast, our approach establishes a fully training-free method that eliminates the need for both large-scale model training and auxiliary driving signals, enabling cross-species motion transfer seamlessly.

\section{Methodology}
\label{sec:method}
In this section, we will introduce the whole pipeline of our proposed system, \methodname. \methodname is a training-free framework via manipulating the attention calculation of the denoising process. To introduce our method, we begin with the introduction of foundational concepts of our base generative framework in \cref{sec:gen}. As our method is not based on the skeleton correspondence, we track the motion flow of subjects in the video playback and build correspondence between the source and target subjects in images (\cref{sec:matching}). To achieve the motion transfer from the source video to the subject, we introduce a novel module, \tpe, for attention manipulation in \cref{sec:tranfer}.

\subsection{Video Generation Framework}
\label{sec:gen}

\paragraph{Diffusion transformer.}
Our framework is built upon WAN-T2V~\cite{wan2025}, which utilizes a DiT architecture integrated with Flow Matching~\cite{lipmanflow}. For efficient processing, an input video is first compressed into a latent space $\mathbf{z} \in \mathbb{R}^{(1+\lfloor\frac{F}{4}\rfloor) \times \frac{H}{8} \times \frac{W}{8} \times C}$ via a 3D causal VAE, where $F$, $H$, and $W$ denote the number of frames, height, and width of the video, respectively, and $C$ represents the latent channel dimension. Unlike traditional Gaussian diffusion, this paradigm models the generative process as a continuous-time probability path where the DiT model predicts a velocity $\mathbf{v}_t$ that transforms initial noise $\mathbf{z}_0$ into the target latent $\mathbf{z}_1$ through linear interpolation $\mathbf{z}_t = t\mathbf{z}_1 + (1 - t)\mathbf{z}_0$. Our \methodname operates within this latent space, intervening in the denoising process by manipulating the spatial-temporal attention maps of the DiT blocks to achieve training-free motion transfer for general subjects. For convenience, we take the simplification of $F \leftarrow 1+\lfloor\frac{F}{4}\rfloor$, $H \leftarrow \frac{H}{8}$, and $W \leftarrow \frac{W}{8}$ to denote the temporal and spatial dimensions of the latent space in the following sections.

\begin{figure}[!t]
    \centering
    \includegraphics[width=\linewidth]{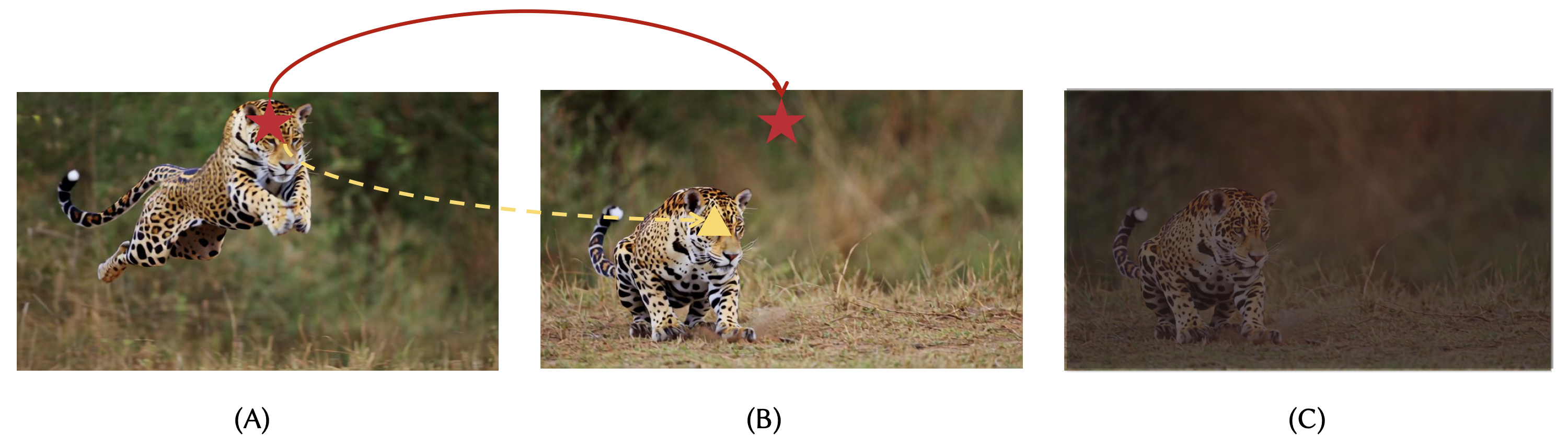}
    \caption{\textbf{Position awareness in self-attention.} We sample two frames from a video. The source point in (A) is marked by a point \textcolor{red}{\ding{72}}. We found the attention weight (overlaid on (C)) of this point \textcolor{red}{\ding{72}} is more aware of the spatial-temporal neighbors \textcolor{red}{\ding{72}}, but not semantically similar ones $\textcolor[RGB]{255, 200, 0}{\blacktriangle}$. }
    \label{fig:relation}
\end{figure}

\paragraph{Self-attention and positional encoding.}
The denoising backbone of WAN consists of $L$ successive transformer blocks, each integrating multi-head self-attention (SA), cross-attention for text conditioning, and a feed-forward network (FFN). Within each SA layer, the input latent $\mathbf{X}$ is first projected into query $\mathbf{Q}$, key $\mathbf{K}$, and value $\mathbf{V}$ tensors, through linear transformations. To capture the complex spatio-temporal dependencies of video data, WAN employs 3D Rotary Positional Embedding (RoPE)~\citep{su2024roformer}. Unlike absolute positional encodings, RoPE injects relative position information by rotating pairs of dimensions in the $\mathbf{Q}$ and $\mathbf{K}$ tensors according to their temporal and spatial coordinates $(f, h, w)$. The attention mechanism then computes the weighted sum of values based on the similarity between queries and keys embedded with positions, which can be formulated as, 
\begin{equation}
    \mathbf{X} \leftarrow \text{Softmax}\left(\frac{\texttt{RoPE}(\mathbf{Q}) \  \texttt{RoPE}(\mathbf{K})^\top}{\sqrt{d}}\right)\mathbf{V},
\end{equation}
where $\mathbf{Q}, \mathbf{K}, \mathbf{V} \in \mathbb{R}^{L \times d} $, $L=F\times H \times W$, and $d$ is the latent dimension of query, key, and value. The resulting attention output $\mathbf{X}$ is subsequently processed by the FFN and fed into the next block. \textit{Particularly}, by injecting positional priors into each latent token, RoPE enables the model to perceive the relative distance between pixels effectively. This mechanism ensures a heightened sensitivity to local structures~\citep{yin2025lazydrag,wang2025characonsist}, as tokens in close proximity exhibit stronger positional correlations during the attention calculation, shown in \cref{fig:relation}. Particularly, points with similar semantics might not have a higher attention weight, as shown in \cref{fig:relation}. This phenomenon reveals the crucial role that positional encoding plays in video generation. Such a relative distance-aware property is crucial for maintaining structural integrity and capturing the nuanced motion dynamics within the spatial-temporal grid.

\subsection{Motion Flows in-and-cross Video Playback}
\label{sec:matching}

\begin{figure}[!t]
    \centering
    \includegraphics[width=\linewidth]{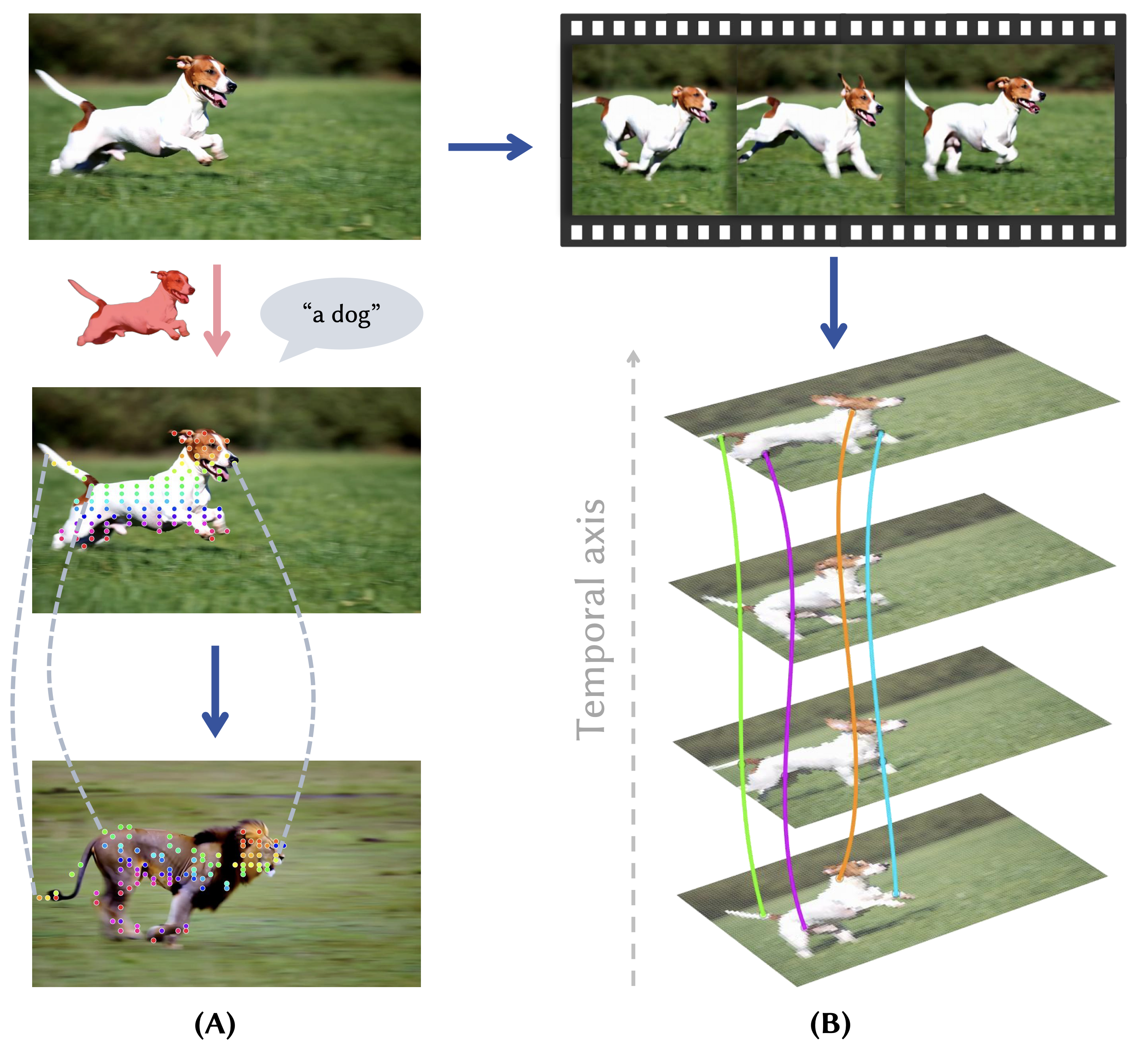}
    \caption{\textbf{Building correspondences across images and video.} 
    (A) Cross-image correspondence: Anchor points \textbf{\textcolor{lightred}{$\mathbf{P}_{src}^1$}} sampled within the source mask are semantically matched to the target subject \textbf{\textcolor{lightblue}{$\mathbf{P}_{tgt}$}} via point matching. 
    (B) Motion flow extraction: The motion flow \textbf{\textcolor{darkred}{$\mathcal{M}^{src}$}} is constructed by tracking the trajectories of these points across the temporal axis of the source video.}
    \label{fig:match}
\end{figure}

\begin{figure*}[!t]
    \centering
    \includegraphics[width=\linewidth]{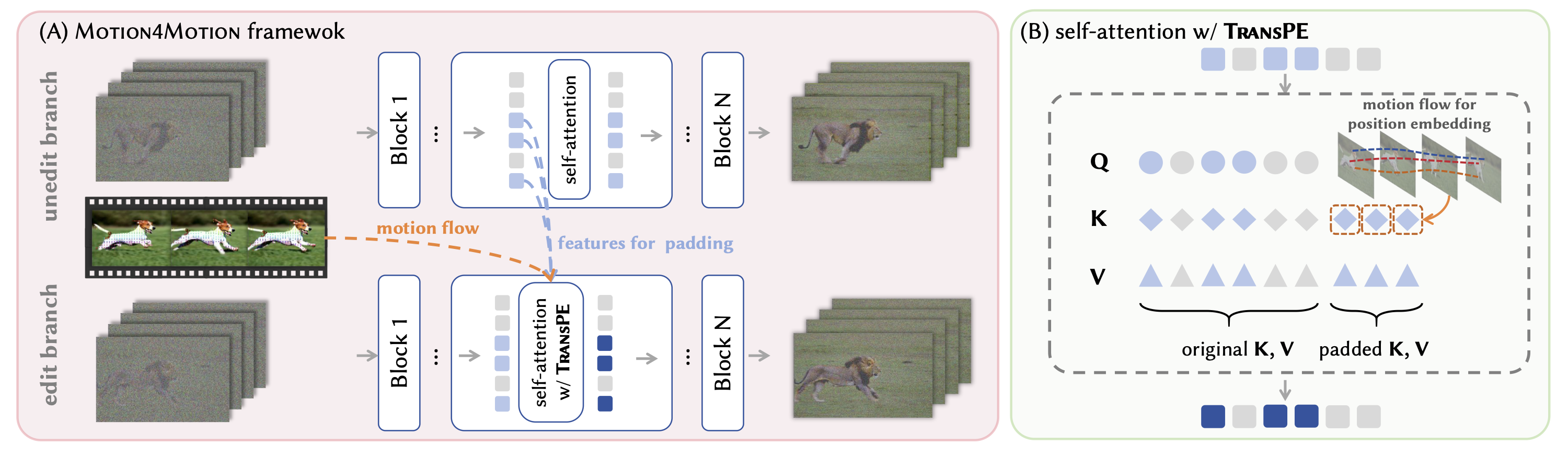}
    \caption{\textbf{System overview of \methodname.} 
    (A) \textbf{Overall Framework.} Our system adds standard self-attention with the \tpe module within DiT blocks to achieve training-free motion transfer. 
    (B) \textbf{Mechanism of \tpe.} The original Query $\mathbf{Q}$ (\protect\scalebox{1.5}{$\bullet$}) keeps unchanged. Key $\mathbf{K}$ (\protect\scalebox{1.5}[0.8]{$\blacklozenge$}) is padded with the appearance key features of the target subject, marked as light blue key tokens \textcolor[RGB]{205,217,239}{\protect\scalebox{1.5}[1]{$\blacklozenge$}}. Additionally, padded keys are also applied to positional encoding using the motion flow trajectories $\mathbf{M}^{tgt}$ from the source video, as marked in the \textcolor[RGB]{210,100,0}{orange dashed boxes}. 
    Value $\mathbf{V}$ (\protect\scalebox{1.2}{$\blacktriangle$}) is augmented by concatenating appearance value features of the target subject, marked as light blue value tokens \textcolor[RGB]{205,217,239}{\protect\scalebox{1.5}[1]{$\blacktriangle$}}. 
    With the proposed pipeline, \methodname faithfully transfers the motion of a source video to a new subject in a another video that follows the source motion. All \textcolor[RGB]{205,217,239}{blue} notations means tokens or features of the ``lion'' in this example. The \textcolor[RGB]{100,140,200}{darker blue} tokens denote the \textit{edited} tokens after \tpe manipulation, in contrast with the \textcolor[RGB]{205,217,239}{lighter blue} tokens representing the original (unedited) target features.}
    \label{fig:system}
\end{figure*}

An essential component in \methodname is establishing the motion flow in the video and then transferring to the target. As shown in \cref{fig:match}, we take an example of transferring the running motion of a dog in a video to a lion. In this stage, we target two essential correspondences, (1) cross-image correspondence between the source and target subjects, and (2) motion flow of the source subjects in the source video. 

\paragraph{Cross-image correspondence.}
To enable motion transfer between subjects with potentially disparate topologies, we first establish a semantic bridge between the source and target subjects in a skeleton-free manner. As shown in \cref{fig:match}-(A), given the first frame of the source video $\mathbf{I}_{src}^1$ and a target subject image $\mathbf{I}_{tgt}$, we employ a subject mask to sample a set of $N$ representative anchor points \textbf{\textcolor{lightred}{$\mathbf{P}_{src}^1 = \{\mathbf{p}_{src, i}^1\}_{i=1}^N$}} on the source character using Grounded SAM-2~\citep{ren2024grounded,ravi2024sam,liu2024grounding}. Technically, the target subject image $\mathbf{I}_{tgt}$ is the first frame of the video generated by unedited prompt. For real target image transfer, we generate its video by an I2V model and then use the inversion technique to obtain the noise. 

We then leverage a semantic matching algorithm (\textit{e.g}., diffusion features~\citep{tang2023emergent}) to find their corresponding coordinates \textbf{\textcolor{lightblue}{$\mathbf{P}_{tgt} = \{\mathbf{p}_{tgt, i}\}_{i=1}^N$}} on the target subject. This point-to-point mapping $\mathcal{C}: \textcolor{lightred}{\mathbf{P}_{src}^1} \to \textcolor{lightblue}{\mathbf{P}_{tgt}}$ serves as the foundation for cross-image transfer, ensuring that the movement of specific semantic parts (\textit{e.g}., the ``legs'' of a dog \textit{vs.} the ``legs'' of a lion) is accurately migrated despite differences in their categories.

\paragraph{Initialization for real image transfer.}
When the target subject is provided as a real image $\mathbf{I}_{tgt}$ rather than generated by the T2V model, we first synthesize a video $\mathbf{V}_{gen}$ from $\mathbf{I}_{tgt}$ using a WAN-I2V model. We then apply inversion~\citep{jiao2025unieditflowunleashinginversionediting} on $\mathbf{V}_{gen}$ to obtain the initial noise latent, which serves as the starting point for the subsequent attention manipulation.

\paragraph{Motion flow in source video.}
While cross-image correspondence handles semantic alignment across subjects, the temporal dynamics must be extracted from the source video playback to drive the animation. As illustrated in \cref{fig:match}-(B), for the sampled anchor points \textbf{\textcolor{lightred}{$\mathbf{P}_{src}^1$}}, we track their trajectories across the subsequent $F-1$ frames using a point tracking framework implemented with DIFT features. This mapping yields a sequence of coordinates for each point, denoted as the motion flow \textbf{\textcolor{darkred}{$\mathcal{M}^{src} = \{\mathbf{P}^f\}_{f=1}^F$}}, where $\mathbf{P}^f$ represents the positions of all anchor points at frame $f$. We mark this mapping as $\mathcal{I}: \textcolor{lightred}{\mathbf{P}_{src}^1} \to \textcolor{lightred}{\mathbf{M}^{src}}$ Specifically, the motion flow (\textit{aka.} trajectory) of the $i$-th point is defined as a set of spatio-temporal coordinates \textbf{\textcolor{black}{$\mathbf{P}_{i}^{f} = \{ (h_i^f, w_i^f) \}_{f=1}^F$}}, illustrated as the motion flow lines in \cref{fig:match}-(B). By decoupling the motion from the source subject into these flow lines, \textbf{\textcolor{darkred}{$\mathbf{M}^{src}$}} provides a topology-agnostic representation of the dynamics, which is subsequently injected into the generative process of the target subject through our attention manipulation in \cref{sec:tranfer}.

Note that the point matching and tracking are conducted within the downsampled coordinate system, downsampled by factors of 4, 8, and 8 along the temporal, height, and width axes, respectively.

\subsection{Video Motion Transfer with \tpe}
\label{sec:tranfer}

Here, we detail how the extracted motion flows are utilized to drive the target subject's generation process. Our strategy consists of two primary steps: mapping the source dynamics to the target subject coordinates and performing a training-free attention manipulation.

\paragraph{Target motion flow construction.}
In \cref{sec:matching}, we introduce the point correspondence across images and the video motion flow. These two mappings are presented as $\mathcal{C}: \textcolor{lightred}{\mathbf{P}_{src}^1} \to \textcolor{lightblue}{\mathbf{P}_{tgt}}$ and $\mathcal{I}: \textcolor{lightred}{\mathbf{P}_{src}^1} \to \textcolor{lightred}{\mathbf{M}^{src}}$, respectively. The core objective of motion transfer is to reproject the points of the target subject onto the spatio-temporal coordinates of the extracted source motion flow, such that $\textcolor{lightblue}{\mathbf{M}^{tgt}} = \textcolor{lightred}{\mathbf{M}^{src}}$. Consequently, the trajectory mapping for the target subject points $\textcolor{lightblue}{\mathbf{P}_{tgt}}$ can be formulated as a composite function: $\mathcal{M} = \mathcal{I} \circ \mathcal{C}^{-1}$. By leveraging this derived mapping $\mathcal{M}: \textcolor{lightblue}{\mathbf{P}_{tgt}} \to \textcolor{lightblue}{\mathbf{M}^{tgt}}$, we establish a relationship between the target subject's semantic points and their intended spatio-temporal destinations in the synthesized video, for inheriting the dynamics of the source.

\paragraph{Attention manipulation with \tpe.}
In practical applications, the target image used for cross-image matching is the first frame of a target subject video. Our goal, therefore, is to reposition the target subject's features to their corresponding locations within the transferred video. This editing process is performed in the latent space of the video via diffusion inversion.
To faithfully migrate the target subject's appearance along the constructed trajectories, we intervene in the self-attention calculation of the DiT blocks. As introduced in the previous text, self-attention calculates the similarity between points with positional encoding. Motivated by this, we introduce a \textsc{Trans}ferring \textsc{P}ositonal \textsc{E}ncoding method (\tpe) to rearrange points of the target subject in editing the attention. 

During the denoising inference, let $\mathbf{Q}, \mathbf{K}, \mathbf{V}$ be the query, key, and value tensors projected from the original noisy latent, \textit{i.e.}, unedited video branch. We introduce the \tpe module to inject the target subject's anchor features into the attention mechanism. Given the latent $\mathbf{K}$ feature of the target subject's first frame $\mathbf{K}_{tgt}^1$ (cached by diffusion inversion), we extract the matched point as $\mathbf{K}^{\star}_{tgt} = \mathbf{K}_{tgt}^1[\textcolor{lightblue}{\mathbf{P}_{tgt}}]$ via the slice operation, which provides the appearance of the target subject. 
We replicate it $F$ times to initialize the sequence of these appearance features $\hat{\mathbf{K}} = [\mathbf{K}^{\star}_{tgt}, \dots, \mathbf{K}^{\star}_{tgt}] \in \mathbb{R}^{F \times H \times W \times d}$. Similarly, we can slice the feature of the subject from the original $\mathbf{V}$ and repeat $F$ times, as $\hat{\mathbf{V}}$, with the same dimension of $\hat{\mathbf{K}}$. 
After that, we re-embed $\hat{\mathbf{K}}$ with positional information using \texttt{RoPE} based on the target motion flow \textbf{\textcolor{lightblue}{$\mathbf{M}^{tgt}$}}. This allows the model to ``look for'' the target subject's features at the newly transferred coordinates. We then augment the original key and value tensors via concatenation, 
\begin{equation}
    \mathbf{K}_{new} = [\texttt{RoPE}(\mathbf{K}), \ \texttt{RoPE}(\hat{\mathbf{K}}, \text{\textcolor{lightblue}{\small$\mathbf{M}^{tgt}$}})], \quad \mathbf{V}_{new} = [\mathbf{V}, \ \hat{\mathbf{V}}],
\end{equation}
while keeping the query $\mathbf{Q}$ unchanged. The updated self-attention operation is then performed as, 
\begin{equation}
    \mathbf{X} \leftarrow \text{Softmax}\left(\frac{\texttt{RoPE}(\mathbf{Q}) \  \mathbf{K}_{new}^\top}{\sqrt{d}}\right)\mathbf{V}_{new}.
\end{equation}
By padding position-aware features into the attention space, \methodname effectively forces the denoising process to synthesize the target subject at the specific coordinates dictated by the motion flow, achieving high-fidelity motion transfer without tuning.

The complete pipeline of \methodname is summarized in \cref{alg:transpe}. Given a source video and target subject, we first extract the motion flow and cross-image correspondence, then perform the denoising process with \tpe attention manipulation within specified step and layer ranges.


\begin{algorithm}[t]
\caption{Motion Transfer with \tpe}
\label{alg:transpe}
\small
\KwIn{Source video $\mathbf{V}_{src}$, target subject $\mathbf{I}_{tgt}$, text prompt $\mathbf{c}$, total steps $T$, total layers $L$, begin/end step $s_b$/$s_e$, begin/end layer $l_b$/$l_e$}
\KwOut{Transferred video $\mathbf{V}_{tgt}$}
\tcp{Stage 1: Motion Flow Extraction}
$\textcolor{lightred}{\mathbf{P}_{src}^1} \leftarrow \texttt{SampleAnchors}(\mathbf{V}_{src}[0])$ \tcp*{anchor points on source first frame}
$\textcolor{lightblue}{\mathbf{P}_{tgt}} \leftarrow \texttt{SemanticMatch}(\textcolor{lightred}{\mathbf{P}_{src}^1}, \mathbf{I}_{tgt})$ \tcp*{cross-image correspondence $\mathcal{C}$}
$\textcolor{darkred}{\mathbf{M}^{src}} \leftarrow \texttt{PointTrack}(\textcolor{lightred}{\mathbf{P}_{src}^1}, \mathbf{V}_{src})$ \tcp*{motion flow in source video $\mathcal{I}$}
$\textcolor{lightblue}{\mathbf{M}^{tgt}} \leftarrow \textcolor{darkred}{\mathbf{M}^{src}}$ \tcp*{target flow via $\mathcal{M} = \mathcal{I} \circ \mathcal{C}^{-1}$}

\tcp{Stage 2: Latent Initialization}
$\mathbf{V}_{gen} \leftarrow \texttt{Generate}(\mathbf{I}_{tgt}, \mathbf{c})$ \tcp*{T2V/I2V generation}
$\mathbf{z}_0 \leftarrow \texttt{Inversion}(\mathbf{V}_{gen})$ \tcp*{deterministic inversion}
Cache $\mathbf{K}_{tgt}^1, \mathbf{V}_{tgt}^1$ from inversion at each layer\;

\tcp{Stage 3: Denoising with \tpe}
\For{$t = 0$ \KwTo $T-1$}{
    \For{$l = 0$ \KwTo $L-1$}{
        Compute $\mathbf{Q}, \mathbf{K}, \mathbf{V}$ from $\mathbf{z}_t$ at layer $l$\;
        \eIf{$t < s_e$ \textbf{and} $l_b \leq l < l_e$}{
            \tcp{Apply \tpe manipulation}
            $\hat{\mathbf{K}} \leftarrow \texttt{Repeat}(\mathbf{K}_{tgt}^1[\textcolor{lightblue}{\mathbf{P}_{tgt}}], F)$\;
            $\hat{\mathbf{V}} \leftarrow \texttt{Repeat}(\mathbf{V}_{tgt}^1[\textcolor{lightblue}{\mathbf{P}_{tgt}}], F)$\;
            $\mathbf{K}_{new} \leftarrow [\texttt{RoPE}(\mathbf{K}),\ \texttt{RoPE}(\hat{\mathbf{K}}, \textcolor{lightblue}{\mathbf{M}^{tgt}})]$\;
            $\mathbf{V}_{new} \leftarrow [\mathbf{V},\ \hat{\mathbf{V}}]$\;
            $\mathbf{X} \leftarrow \texttt{Softmax}\left(\frac{\texttt{RoPE}(\mathbf{Q}) \cdot \mathbf{K}_{new}^\top}{\sqrt{d}}\right)\mathbf{V}_{new}$\;
        }{
            $\mathbf{X} \leftarrow$ standard self-attention\;
        }
    }
    $\mathbf{z}_{t+1} \leftarrow$ update with velocity prediction\;
}
$\mathbf{V}_{tgt} \leftarrow \texttt{Decode}(\mathbf{z}_T)$\;
\end{algorithm}
\section{Experiments}
\label{sec:experiment}

\subsection{Setting}
\label{sec:expset}

\subsubsection{Implementation details} Our method is implemented on top of the WAN-14B-T2V~\citep{wan2025} (480p resolution) model. By default, attention manipulation is applied across layers [0, 40] until step 35 out of 50 denoising steps. \tpe is applied in all self-attention layers within this range. Point matching is performed via diffusion feature matching, while subject segmentation is obtained using SAM-2~\citep{ravi2024sam}. To align the segmentation masks and point coordinates with the latent space resolution, both are downsampled by a factor of $1/8$. All experiments are conducted on one NVIDIA H-800 GPU. 

\subsubsection{Baselines} Most existing methods in this field focus on human or human-like motion transfer and typically rely on predefined global motion signals or explicit pose representations. For example, approaches such as WAN-animate~\citep{cheng2025wan} heavily depend on human pose detectors, whose performance may be unstable in non-human-like character scenarios.
In this paper, to evaluate general fine-grained character motion control, we first adopt two state-of-the-art methods, FlexAct~\cite{zhang2025flexiact} and MotionClone~\citep{lingmotionclone}, as primary baselines. We further compare our approach with global motion transfer methods, including MotionDirector~\cite{zhao2024motiondirector} and RoPECraft~\cite{gokmen2025ropecraft}. Additionally, we compare with recent training-required trajectory-based methods, Diffusion-As-Shader~\cite{gu2025diffusionshader} and WAN-Move~\cite{chu2025wanmove}, which train dedicated motion control modules on large-scale trajectory data. Since the implementation of the recent method MotionV2V~\citep{burgert2025motionv2veditingmotionvideo} and MotionShot~\citep{liu2025motionshot} was not publicly available before submission, we discuss them in related work.

\subsubsection{Benchmark and evaluation protocol}
We evaluate \methodname on the established animal (33 pairs) and human motion (123 pairs) transfer benchmarks introduced by~\citet{zhang2025flexiact}. Since our framework is built upon the WAN-T2V-14B foundation model, we adapt the evaluation protocol by first performing a deterministic inversion~\citep{jiao2025unieditflowunleashinginversionediting} of the source video to obtain the base latent path, followed by our attention manipulation to synthesize the transferred results. To quantify the quality of the transferred results, we employ four standard metrics:
(1) \textit{Textual Similarity} (TS): calculated via CLIP~\citep{radford2021learning} to measure the semantic consistency between the generated frames and the prompt. 
(2) \textit{Motion Fidelity} (MF): which utilizes tracklets computed by Co-tracker~\cite{karaev2025cotracker3} to measure the similarity between motion trajectories in unaligned videos.
(3) \textit{Temporal Consistency} (TC): quantified by the average CLIP image feature similarity between all frame pairs to ensure smoothness and coherence. 
(4) \textit{Appearance Consistency} (AC): which reflects the identity preservation by calculating the average CLIP~\cite{radford2021learning} similarity between the target image and the generated video frames.
Additionally, to evaluate the fine-grained pose alignment between the source and target motion, we introduce a new benchmark with 50 image-video pairs with different animals. Specifically, the category of these examples can be covered by a cross-category pose detector. Consequently, we evaluate the \textit{pose similarity (PS)} between the source and target motions with a precise detector~\citep{yang2024x}. 

\label{sec:withbaseline}

\begin{table*}[htbp]
    \centering
    \caption{\textbf{Main quantitative comparison with baselines.}}
    \label{tab:results}
    \resizebox{\textwidth}{!}{\setlength{\tabcolsep}{6pt}{
        \begin{tabular}{l|cccc|cccc|ccccc}
            \toprule
            \multirow{3}{*}{{Method}} & \multicolumn{4}{c|}{Human} & \multicolumn{4}{c|}{Animal} & \multicolumn{5}{c}{Ours} \\
            \cmidrule(lr){2-5} \cmidrule(lr){6-9} \cmidrule(lr){10-14}
            & {TS} $\uparrow$ & {MF} $\uparrow$ & {TC} $\uparrow$ & {AC} $\uparrow$ 
            & {TS} $\uparrow$ & {MF} $\uparrow$ & {TC} $\uparrow$ & {AC} $\uparrow$ 
            & {TS} $\uparrow$ & {MF} $\uparrow$ & {TC} $\uparrow$ & {AC} $\uparrow$ & {PS} $\uparrow$ \\
            \midrule
            MotionDirector~\citeyearpar{zhao2024motiondirector}  & 0.255 & 0.312 & 0.915 & 0.887 & 0.248 & 0.298 & 0.902 & 0.875 & 0.251 & 0.305 & 0.908 & 0.881 & 0.342 \\
            RoPECraft~\citeyearpar{gokmen2025ropecraft}       & 0.241 & 0.330 & 0.907 & 0.894 & 0.235 & 0.315 & 0.895 & 0.882 & 0.238 & 0.322 & 0.901 & 0.888 & 0.355 \\
            MotionClone~\citeyearpar{lingmotionclone}     & 0.258 & 0.381 & 0.937 & 0.900 & 0.252 & 0.365 & 0.924 & 0.891 & 0.255 & 0.373 & 0.931 & 0.896 & 0.408 \\
            FlexiAct~\citeyearpar{zhang2025flexiact}         & 0.269 & 0.391 & 0.928 & 0.945 & 0.261 & 0.378 & 0.915 & 0.932 & 0.265 & 0.384 & 0.922 & 0.938 & 0.415 \\
            \midrule
            \methodname          & \textbf{0.288} & \textbf{0.452} & \textbf{0.955} & \textbf{0.971} & \textbf{0.282} & \textbf{0.441} & \textbf{0.948} & \textbf{0.962} & \textbf{0.285} & \textbf{0.448} & \textbf{0.952} & \textbf{0.967} & \textbf{0.543} \\            \bottomrule
        \end{tabular}
    }
    }
\end{table*}

\subsection{Evaluation}

\subsubsection{Quantitative evaluation} We compare our method with several state-of-the-art baselines in \cref{tab:results}, where \methodname consistently achieves the best performance across all metrics. The superiority of our approach can be attributed to its fine-grained control mechanism. Compared to earlier frameworks like MotionDirector and RoPECraft, which primarily rely on global representations or model-level tuning, \methodname operates at a point-to-pixel level. This enables our model to capture local motion nuances that are typically lost in holistic control paradigms, leading to significant gains in Motion Fidelity (MF). Furthermore, \methodname outperforms recent mask-based methods such as MotionClone and FlexiAct. While these approaches improve motion quality by constraining attention within the subject's silhouette, they lack explicit semantic guidance between the source and target. In contrast, our point-to-point mapping establishes a direct semantic bridge, ensuring that each part of the target subject follows the intended trajectory with high precision. This is reflected in the substantial improvement in Pose Similarity (PS), demonstrating our method's robustness in handling complex character transfers.

\begin{figure*}[!t]
    \centering
    \includegraphics[width=\linewidth]{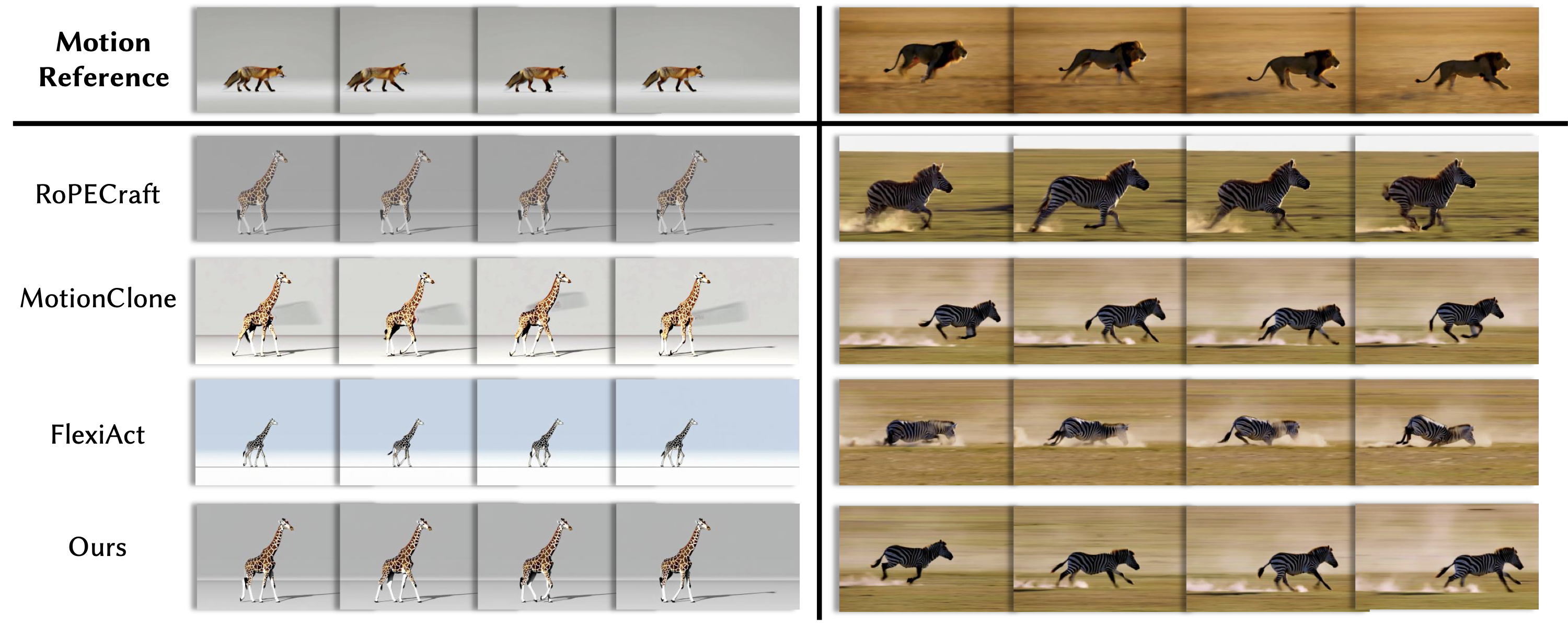}
    \caption{\textbf{Comparison of \methodname with baselines.} We focus on the preservation of pose details across individual frames. Most baseline methods exhibit varying degrees of visual artifacts or appearance drifting. In contrast, our method maintains high-fidelity pose alignment and appearance consistency.}
    \label{fig:mainres}
\end{figure*}

\subsubsection{Qualitative evaluation}
\begin{figure*}[!t]
    \centering
    \includegraphics[width=0.98\linewidth]{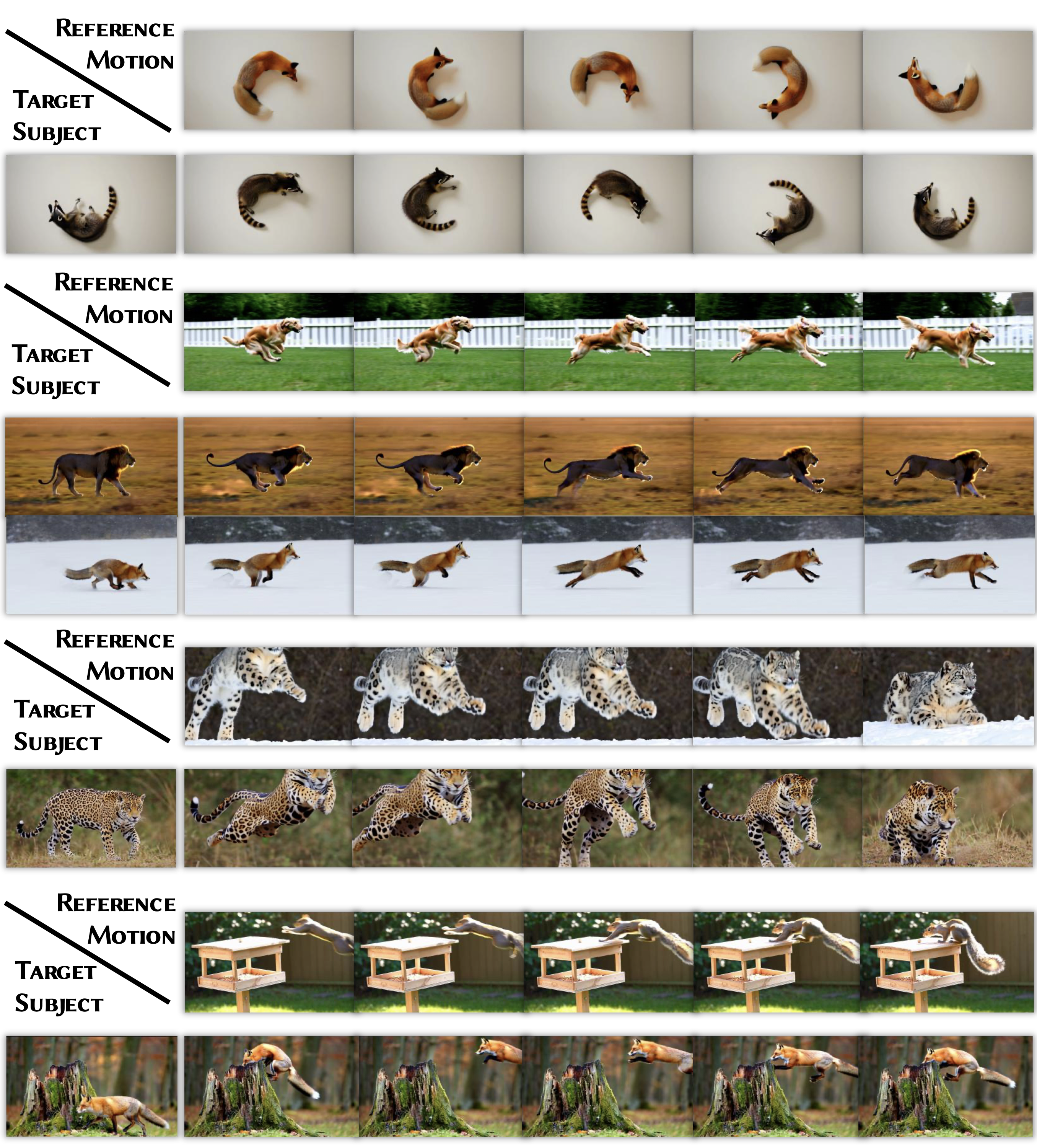}
    \caption{\textbf{Examples of the motion transfer results bt \methodname.} }
    \label{fig:moreres}
\end{figure*}

We present a visual comparison between our method and several state-of-the-art baselines in Fig.~\ref{fig:mainres}, focusing on challenging cross-species motion transfer tasks (\textit{e.g.}, transferring a fox's gait to a giraffe, and a lion's pounce to a zebra). As illustrated, existing baselines struggle to maintain a balance between motion fidelity and visual quality. Specifically, RoPECraft~\citep{gokmen2025ropecraft} often produces stiff movements and fails to capture the fluid leg dynamics of the reference. FlexiAct~\citep{zhang2025flexiact} suffers from significant identity distortion and texture blurring, particularly around the limbs (as seen in the giraffe's shadow and the zebra's legs), failing to preserve the structural integrity of the target animal. While MotionClone~\citep{lingmotionclone} achieves better motion magnitude, it introduces noticeable \textit{ghosting artifacts} and \textit{temporal appearance drifting}.
A key limitation of these baselines is their over-reliance on global motion features or their lack of explicit structural guidance, which makes precise \textbf{fine-grained pose alignment} extremely difficult. In contrast, our approach achieves superior semantic pose alignment even across disparate morphologies. Despite the vast differences in limb proportions between a fox and a giraffe, our method precisely retargets the reference gait while maintaining high-fidelity appearance and sharp textures without the need for skeletal priors or fine-tuning. More results are in \cref{fig:moreres}.

\subsubsection{Real-image/video evaluation} 
\begin{figure*}[!t]
    \centering
    \begin{minipage}{0.35\linewidth}
        \centering
        \includegraphics[width=0.85\linewidth]{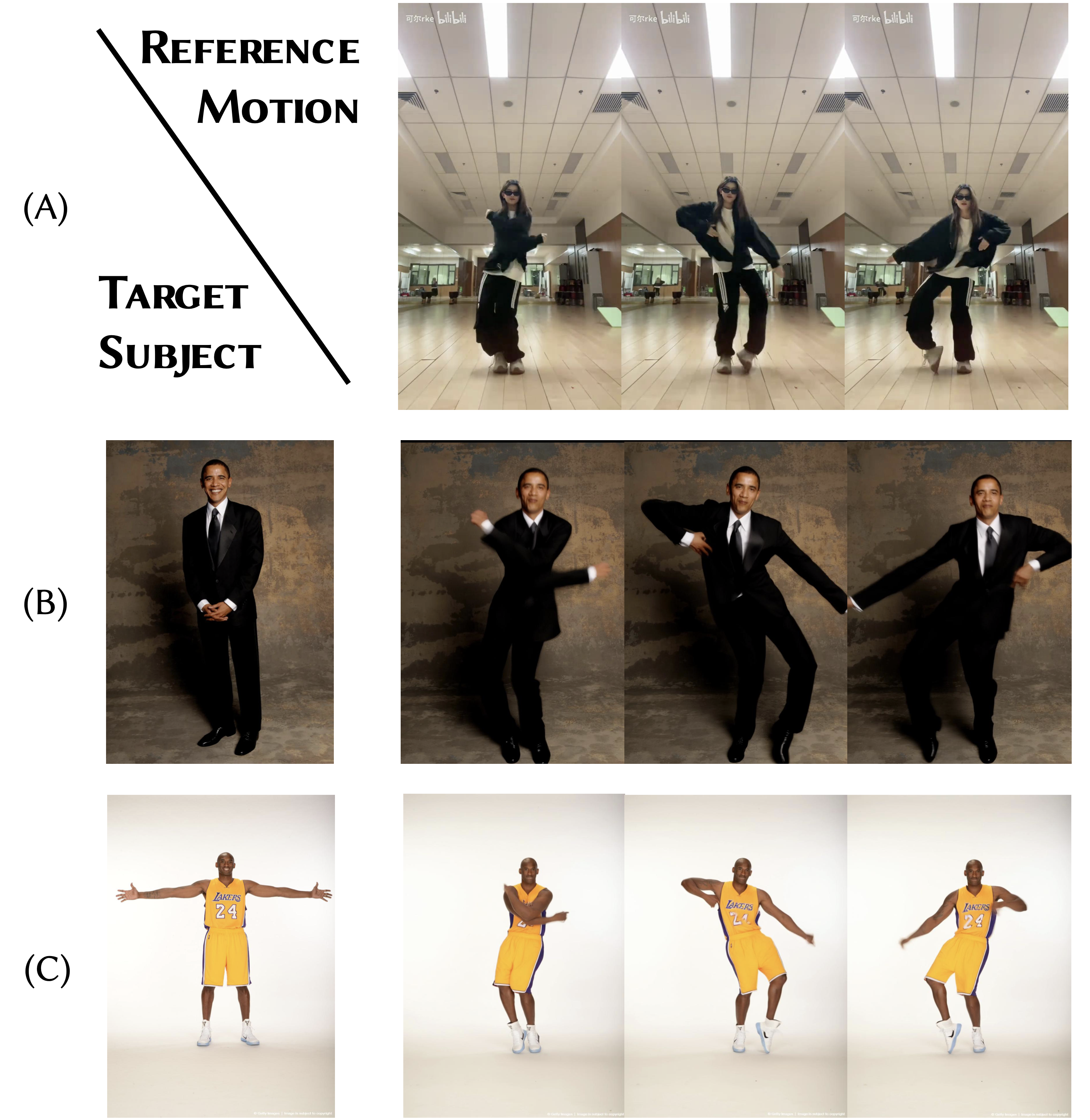}
        \caption{\textbf{Real image motion motion transfer \methodname.} The image of the driven subject and the motion driving video are from the Internet. The transfer results show the good performance of \methodname.}
        \label{fig:real-img}
    \end{minipage}
    \hfill 
    \begin{minipage}{0.63\linewidth}
        \centering
        \includegraphics[width=0.95\linewidth]{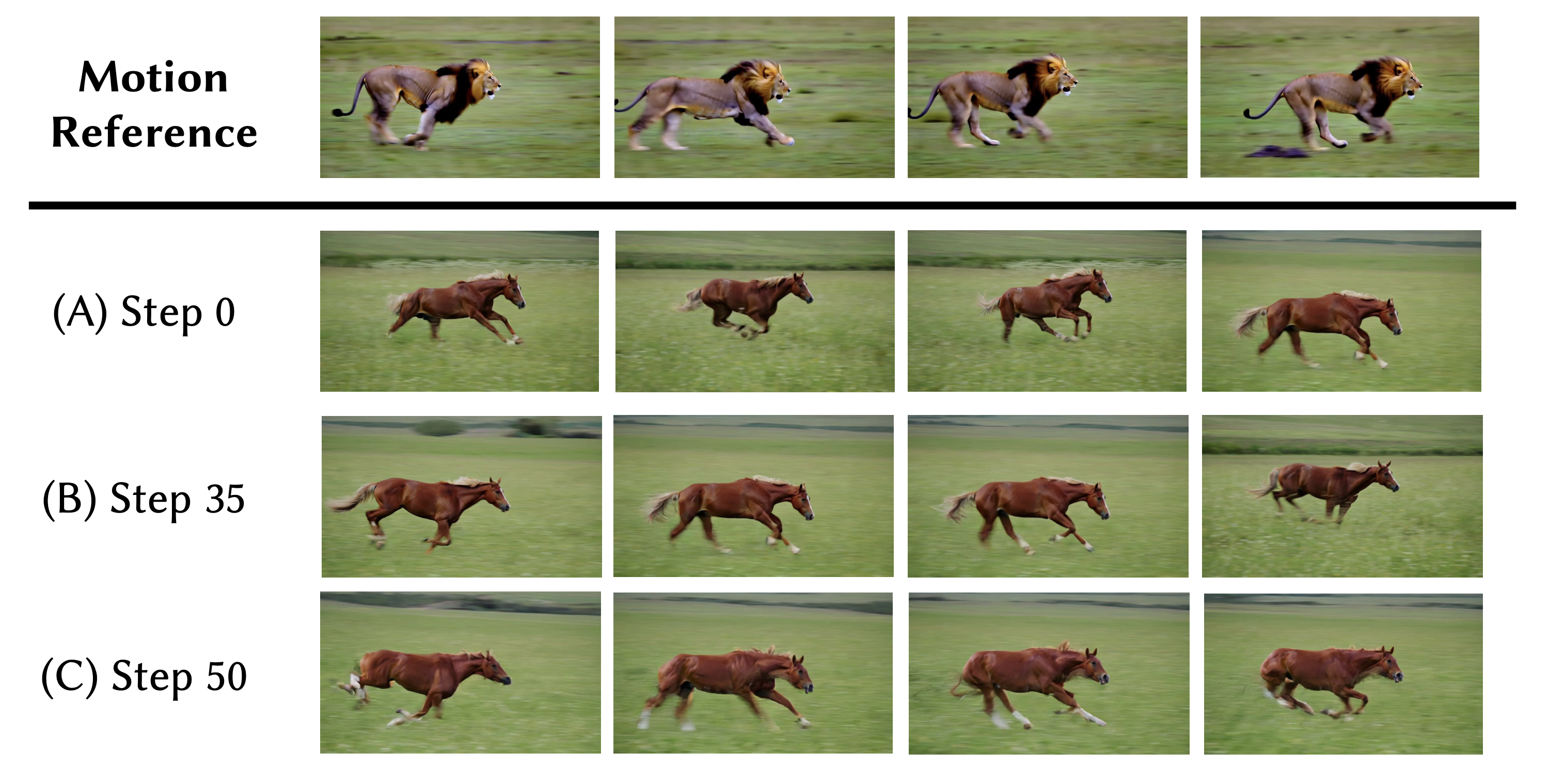}
        \caption{\textbf{Ablation on manipulation steps.} (A) Without manipulation (Step 0), the model fails to relocate target features along the intended trajectories. (B) Our default strategy (Step 35) achieves an optimal balance between motion fidelity and visual realism. (C) Full-step manipulation (Step 50) causes semantic confusion, resulting in texture stretching and artifacts.}
        \label{fig:ablation-step}
    \end{minipage}
\end{figure*}

To further evaluate the robustness and generalization capabilities of \methodname, we perform motion transfer using ``in-the-wild'' images and videos sourced from the Internet. As shown in Fig.~\ref{fig:real-img}, we animate static portraits of diverse subjects, including public figures with distinct stylistic features, using a complex dance sequence as the source motion. Despite our method being without any training, \methodname successfully retargets the dynamic motion while faithfully preserving the subjects' identity, clothing textures, and structural integrity. This zero-shot performance on diverse real-world data demonstrates that our approach does not overfit to specific datasets and can handle diverse in-the-wild scenarios without additional fine-tuning.

\subsection{User Study}

We conducted a subjective evaluation to assess the perceptual quality of the generated videos. 
The study followed a blind pairwise comparison protocol~\cite{zhang2025flexiact,zhao2024motiondirector}, where participants were presented with two videos side-by-side: one generated by the Base Model (WAN-I2V-14B) and the other by a competing method (or ours). 
For each of the 50 randomly selected test cases, all 10 raters were asked to select the preferred result based on two specific criteria: \textit{Motion Consistency} (how accurately the transferred motion matches the source motion) and \textit{Appearance Consistency} (how well the visual identity of the subject is preserved).
As reported in \cref{tab:human_eval}, our method demonstrates a significant advantage over the baseline, on both motion and appearance consistency. 
These results confirm that our approach not only ensures faithful motion transfer but also maintains higher visual fidelity compared to baselines.

\begin{table}[t]
    \centering
    \caption{\textbf{Human evaluation results compared to others.} Our method outperforms baselines across all dimensions. The ``Base Model'' refers to WAN-I2V-14B with \tpe (K-V concatenation) applied, which serves as the anchor for pairwise comparison.}
    \label{tab:human_eval}
    \begin{tabular}{lcc}
        \toprule
        & \textbf{Motion} & \textbf{Appearance} \\
        \textit{v.s.} Base Model& \textbf{Consistency} & \textbf{Consistency} \\
        \midrule
        MotionDirector  & 44.8 \textit{v.s.} 55.2 & 49.7 \textit{v.s.} 50.3 \\
        RoPECraft       & 67.1 \textit{v.s.} 32.9 & 62.6 \textit{v.s.} 37.4 \\
        \midrule
        MotionClone     & 71.7 \textit{v.s.} 28.3 & 68.6 \textit{v.s.} 31.4 \\
        FlexAct         & 78.4 \textit{v.s.} 21.6 & 62.1 \textit{v.s.} 37.9 \\
        \midrule
        \methodname     & \textbf{92.5} \textit{v.s.} \textbf{7.5} & \textbf{87.8} \textit{v.s.} \textbf{12.2} \\
        \bottomrule
    \end{tabular}
\end{table}

\section{Application: Teaching a Table Walking}
\label{sec:app}

\begin{figure*}[!h]
    \centering
    \includegraphics[width=0.95\linewidth]{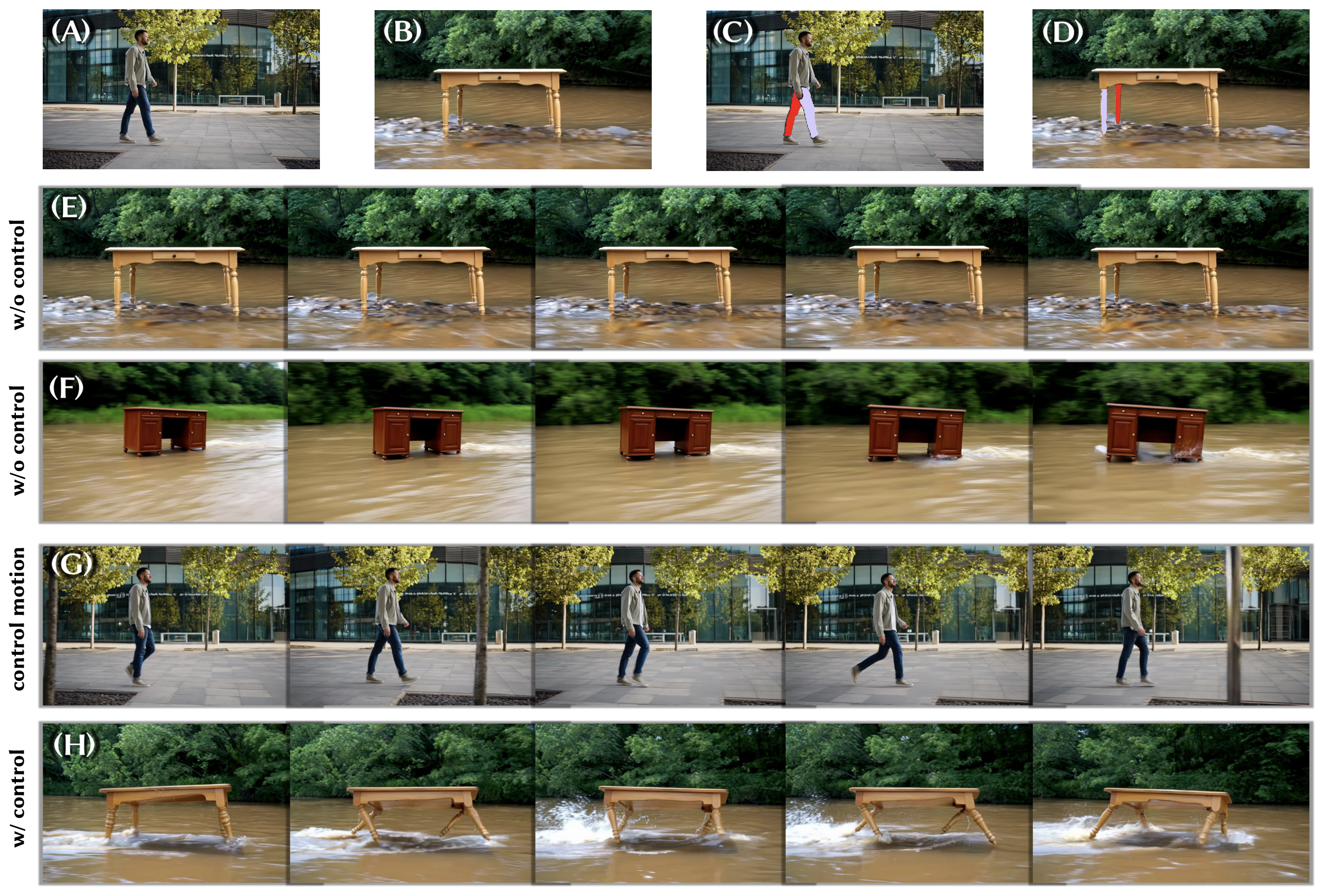}
    \caption{\textbf{Application of \methodname in Cross-domain Motion Synthesis.} We empower a vanilla T2V model with the capability to animate inanimate objects using biological motion trajectories. Given a reference video of a walking person (A) and a target static image of a table (B), our method extracts the motion semantics (G) and applies them to the target. While baseline approaches (WAN-I2V-14B, WAN-T2V-14B) without our control (E, F) produce either static motions or simple sliding motions. With the binding legs from a man to two specified legs in (C) and (D), \methodname (H) successfully retargets the walking gait onto the table's legs while maintaining structural integrity and environmental consistency.}
    \label{fig:app}
\end{figure*}

Although a large video generation model shows impressive generation ability in various scenarios, it still struggles with some novel concept composition~\cite{shi2025imbalance}, such as ``\texttt{A desk coming to life, running rapidly along a muddy riverside.}''. As the \textbf{novel concept composition} of ``\texttt{running}'' and ``\texttt{desk}'' differs substantially, the T2V model is quite hard to compose a reasonable result for the model. As shown in \cref{fig:app}(E-F), the model always synthesizes the static desk motions or generates a sliding motion, due to the limited imagination ability of the base model. 

To address this, we introduce an external source to control the motion. We explore the boundaries of our algorithm by performing \textbf{cross-morphology transfer}: driving a static table (\cref{fig:app}B) using a video of a walking human (\cref{fig:app}A). This presents a significant challenge due to the distinct structural differences. To render this problem tractable, inspired by ``bone binding'' in 3D animation~\citep{rokoko,autorigpro}, we explicitly bind the human legs to the table's legs using SAM2 masks, as shown in \cref{fig:app}(C-D). Consequently, \methodname constrains the diffusion feature matching within these masked regions, mitigating the structural ambiguity caused by morphological disparities and successfully transferring the gait to the table (\cref{fig:app}H), synced with the motion in source video (\cref{fig:app}G).

\section{Conclusion and Discussion}
\label{sec:conclusion}

In this work, we presented \methodname, a novel training-free framework that achieves high-fidelity motion transfer across diverse subjects without relying on skeletal priors or specific fine-tuning. By leveraging the proposed \tpe module, our approach effectively injects motion flows from a source video into the self-attention mechanism of a pre-trained Diffusion Transformer, enabling precise semantic alignment even across disparate morphologies. Extensive evaluations demonstrate that \methodname significantly outperforms state-of-the-art baselines in both motion fidelity and appearance preservation. Furthermore, the ability to animate inanimate objects (\textit{e.g.}, a walking table) highlights the generalization capability of our method. We believe this work offers a flexible and efficient solution for video generation, opening new avenues for creative animation workflows.

\begin{acks}

The author team of \methodname would like to convey sincere appreciation to all reviewers and committee members for their significant efforts in enhancing the quality of this work. We sincerely acknowledge Bohong Chen, Yukai Shi, and Shiyi Zhang for their thoughtful suggestions and discussions on this paper. 
\end{acks}

\clearpage

\clearpage
\bibliographystyle{ACM-Reference-Format}
\bibliography{reference}



\end{document}